\documentclass[runningheads]{llncs}

\usepackage[mobile]{eccv}

\usepackage{eccvabbrv}

\usepackage{graphicx}
\usepackage{booktabs}
\usepackage{soul}

\usepackage[accsupp]{axessibility}  

\usepackage[dvipsnames]{xcolor}
\usepackage{multirow}

\usepackage{pifont}
\usepackage{float}

\newcommand{\cmark}{\ding{51}}%
\newcommand{\xmark}{\ding{55}}%

\usepackage[most]{tcolorbox}
\usepackage{colortbl} 
\usepackage{cuted}
\usepackage{listings}
\tcbuselibrary{listingsutf8}
\usepackage{fvextra}
\newcommand{\na}{--}

\newcommand{\eb}[1]{$\mathbf{{#1}}$}
\newcommand{\eu}[1]{$\underline{{#1}}$}

\newcommand{\ours}{PhysMoDPO\xspace}%

\definecolor{aliceblue}{rgb}{0.94, 0.97, 1.0}

\usepackage[subtle]{savetrees} 

\usepackage[pagebackref,breaklinks,colorlinks,citecolor=eccvblue]{hyperref}

\usepackage{orcidlink}

\begin{document}


\title{PhysMoDPO: Physically-Plausible Humanoid Motion with Preference Optimization}

\titlerunning{PhysMoDPO}

\author{Yangsong Zhang$^{1}$\and
Anujith Muraleedharan$^{1}$\and
Rikhat Akizhanov$^{1}$\and
Abdul Ahad Butt$^{1}$\and
Gül Varol$^{2}$\and
Pascal Fua$^{3}$\and
Fabio Pizzati$^{1}$\and
Ivan Laptev$^{1}$
}

\authorrunning{Y.~Zhang et al.}

\institute{Mohamed Bin Zayed University of Artificial Intelligence (MBZUAI) \and
LIGM, École des Ponts, IP Paris, Univ Gustave Eiffel, CNRS \and
École polytechnique fédérale de Lausanne (EPFL)\\
\url{https://mael-zys.github.io/PhysMoDPO/}}

\maketitle

\begin{abstract}
Recent progress in text-conditioned human motion generation has been largely driven by diffusion models trained on large-scale  human motion data. 
Building on this progress, recent methods attempt to transfer such models for character animation and real robot control by applying a Whole-Body Controller (WBC) that converts diffusion-generated motions into executable trajectories.
While WBC trajectories become compliant with physics, they may expose substantial deviations from original motion.
To address this issue, we here propose \ours, a Direct Preference Optimization framework.
Unlike prior work that relies on hand-crafted physics-aware heuristics such as foot-sliding penalties, we integrate WBC into our training pipeline and optimize diffusion model such that the output of WBC becomes compliant both with physics and original text instructions. 
To train \ours we deploy physics-based and task-specific rewards and use them to assign preference to synthesized trajectories.
Our extensive experiments on text-to-motion and spatial control tasks demonstrate consistent improvements of \ours in both physical realism and task-related metrics on simulated robots. Moreover, we demonstrate that \ours results in significant improvements when applied to zero-shot motion transfer in simulation and for real-world deployment on a G1 humanoid robot. 
\keywords{Human motion synthesis \and Post-training \and Embodied AI \and Robotics}
\end{abstract}

\section{Introduction}
\label{sec:intro}

\begin{figure}[tb]
  \centering
   \includegraphics[width=0.99\textwidth,trim={0.75cm 9cm 7.2cm 2cm},clip]{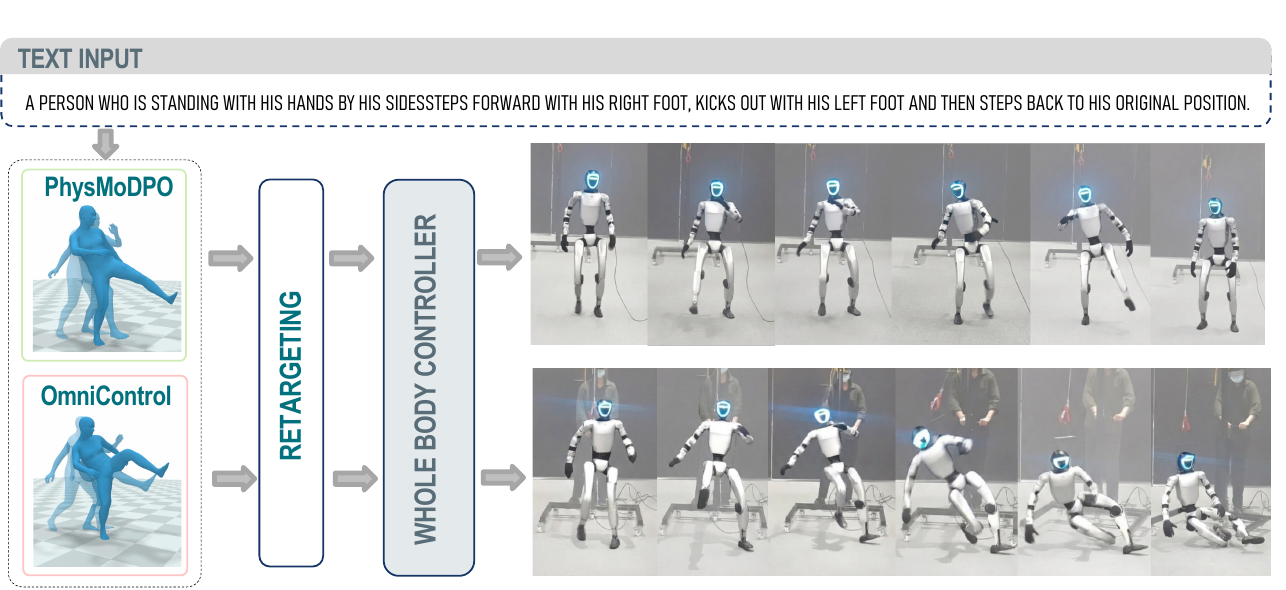}

  \caption{
  \textbf{PhysMoDPO generates motions that follow textual instructions while respecting physical constraints.} Compared to prior methods, our approach produces motions that remain stable and physically realistic when deployed on the Unitree G1 robot.
  }
  \label{fig:teaser}
\end{figure}

The progress of diffusion models has largely advanced the text driven motion generation. Given an input condition (\eg, a text description or additional spatial constraints), a model synthesizes a human motion sequence, enabling applications for animation, virtual avatars, and gaming~\cite{tevet2022MDM,guo2023momask,xie2023omnicontrol,xiao2025motionstreamer}. Among these applications, embodied downstream tasks, especially humanoid robotics, are particularly compelling: if a generative model could reliably produce physically consistent, controllable motions, it could provide a scalable source of reference behaviors for motion tracking and policy training.

However, deploying generated motions in a physics-based setting exposes a key limitation of current generators. Diffusion models are typically trained and evaluated in a kinematic space, where performance is judged by distributional similarity and alignment with the condition signal. In contrast, robotics requires motions that remain feasible under dynamics and contacts: feet should not skate, the center of mass must stay supported.
In practice, when a generated motion is executed via a Whole-Body Controller (WBC) in simulation, a standard strategy in humanoid control~\cite{peng2018deepmimic,peng2021amp,liao2025beyondmimic}, 
the controller may significantly alter the implausible motion to satisfy constraints. 
This introduces a discrepancy between what the generator produces and what is actually realized in the simulation. 
As a result, a model that performs well under kinematic metrics may still be unreliable for embodied deployment.

To allow plausible human motion for deployment in real robotics embodiments, current efforts toward physically plausible motion modeling follow two main directions. 
One line injects physics priors through test-time hand-crafted losses or auxiliary refinement modules~\cite{li2019estimating,rempe2020contact,zhang2024physpt,zhang2025plug}. While effective in specific cases, such approaches can shift the output distribution to adapt physics constraints and thus may largely harm the task-specific performance. 
Another line optimizes generators with heuristics rewards via reinforcement learning~\cite{han2025reindiffuse,wen2025hy}. Although promising and visually plausible, the performance highly depends on the heuristics functions, \ie, floating, and sliding, which makes it difficult to measure complex physical dynamics. Therefore, directly scaling hand-crafted rewards to the complicated real-world scenario is non-trivial.

In this work, we propose \ours, a physics-guided post-training framework for diffusion-based motion generators. 
Specifically, we integrate WBC into our training pipeline to measure how close the motion is to transfer to executable trajectory. In this way, the reward could directly cover the physical aspects such as dynamics, floating and jittering.
As for our finetuning framework, inspired by DPO~\cite{rafailov2023direct} alignment in image and video generation~\cite{wallace2024diffusion,liu2025videodpo}, we automatically construct preference data through pretrained WBC.
Concretely, for each condition, we exploit stochastic sampling to generate multiple candidate motions, execute each candidate through the WBC, DeepMimic~\cite{peng2018deepmimic}, to obtain simulated motions, and then compute (i) physics-based rewards that reflect trackability and contact realism for robots, and (ii) task-specific rewards that measure whether the tracked motion still matches the input condition. Importantly, computing task rewards after tracking directly targets the evaluation mismatch: we optimize the generator for motions that remain both physically feasible and condition-faithful. 

Extensive experiments on text-to-motion and spatial control tasks show that \ours consistently improves physical realism while preserving and often improving task-related metrics on simulated robots. Moreover, we retarget and deploy the generated motions to real robots in a zero-shot way, i.e., without additional motion refinement. We demonstrate the zero-shot generalization to simulation and real-world deployment on Unitree G1, suggesting that post-training a generator with physics-guided preferences can produce motions that transfer beyond kinematic benchmarks.

\paragraph{Contributions.}
In summary, our contributions are three-fold:
(1) We propose \ours, an automatic generate--finetune pipeline that leverages a pretrained WBC to post-train diffusion motion generators toward robotics-oriented physical consistency.
(2) We design physics-based rewards that measure trackability and contact realism, together with task-specific rewards that evaluate condition fidelity after tracking policy, enabling preference supervision aligned with embodied deployment.
(3) We conduct extensive experiments on text-to-motion and spatial control, demonstrating improved physical realism and task-related metrics in simulation, and showing zero-shot transfer to Unitree G1 in both simulation and real-world deployment.

\section{Related Works}
\label{sec:related_work}

\noindent\textbf{Controllable human motion generation.}
Human motion generation has been studied under various conditioning signals, including action labels~\cite{petrovich21actor}, speech audio~\cite{li2021audio2gestures,zhu2023taming}, music~\cite{li2022danceformer,tseng2023edge}, text descriptions~\cite{guo2023momask,tevet2022MDM,zhang2023generating,petrovich22temos}.
More recently, in the domain of text-to-motion generation, research has shifted towards more fine-grained controllable motion synthesis, exploring temporal or spatial composition~\cite{TEACH:3DV:2022,SINC:ICCV:2023,petrovich2024multi}, streaming generation~\cite{xiao2025motionstreamer}, integrating scene context~\cite{wang2022humanise,zhao2023synthesizing}, or object~\cite{li2023object,li2023controllable} into the generation framework and detailed spatial joint control~\cite{karunratanakul2023guided,xie2023omnicontrol,dai2024motionlcm,pinyoanuntapong2025maskcontrol}.
Despite this progress, generated motions often fail to execute in physics-based simulators, exhibiting artifacts such as the center of mass that lies outside the body support.
To address this gap, we integrate a pre-trained physics-based method to compute physics-based rewards as part of learning signals for DPO~\cite{rafailov2023direct} fine-tuning, thereby improving the physical plausibility.\smallskip

\noindent\textbf{Physically plausible human motion modeling.}
As shown in  Table~\ref{tab:compare_rw}, one direction in physically plausible motion modeling is to apply hand-craft optimization~\cite{li2019estimating,rempe2020contact} or additional models~\cite{zhang2024physpt,zhang2025plug,tevet2024closd} to project the predicted noisy human motions. Previous works~\cite{li2019estimating,rempe2020contact} typically apply optimization through the Euler-Lagrange equations. PhysPT~\cite{zhang2024physpt} further proposes a physics-aware transformer through self-supervised learning with Euler-Lagrange equations. Recent works~\cite{zhang2025plug,li2025morph,yuan2023physdiff} further use physics-based methods to refine motion to avoid floating and sliding problems. 
However, simply applying physics-based methods may possibly change the output motion distribution, which may require test-time finetuning. Another branch of work directly refines motion generators with extra physics-based rewards~\cite{han2025reindiffuse,wen2025hy} or models~\cite{li2025morph}. ReinDiffuse~\cite{han2025reindiffuse} and HY-Motion 1.0~\cite{wen2025hy} define some hand-crafted rewards such as floating and sliding, and then apply PPO~\cite{schulman2017proximal} and GRPO~\cite{shao2024deepseekmath} to refine the motion generators, respectively. Though effective, those hand-crafted rewards can hardly cover other essential aspects such as abnormal mass center. To address this, Morph~\cite{li2025morph} trains a physics-based model to refine the generated motion and finetunes the motion generator with the refined data. However, when the motion from generators is too noisy, this training strategy could possibly harm the physics-based model. In this work, we use a pre-trained physics-based model to solely compute the physics rewards and then apply DPO~\cite{rafailov2023direct} finetuning on the motion generator.\smallskip

\begin{table}[tb]
  \caption{\textbf{Comparison with previous work.}
  We compare alternative ways of incorporating physics constraints into human motion generation.
  In contrast to other methods, \ours make use of dynamics-aware reward and does not require test-time optimization nor additional trainable modules.
  }
  \label{tab:compare_rw}
  \centering
  \setlength{\tabcolsep}{5pt}
  \small
  \resizebox{0.99\textwidth}{!}{
  \begin{tabular}{@{}l l|cc|c|c|c@{}}
    \toprule
    Category & Method &
    \begin{tabular}[c]{@{}c@{}}Approach\end{tabular} &
    \begin{tabular}[c]{@{}c@{}}Applied\\ at\end{tabular} &
    \begin{tabular}[c]{@{}c@{}}No test-time\\ opt.\end{tabular} &
    \begin{tabular}[c]{@{}c@{}}No Extra trainable\\ module\end{tabular} &
    \begin{tabular}[c]{@{}c@{}}Dynamics-aware\\ reward\end{tabular} \\
    \midrule

    \multirow{5}{*}{\begin{tabular}[c]{@{}l@{}}\emph{Constraints}\\[-1pt]\emph{/ Projection}\end{tabular}}
      & Li et~al.~\cite{li2019estimating}         & Optimization   & Train/Infer & \xmark & \cmark & \na \\
      & Rempe et~al.~\cite{rempe2020contact}      & Optimization   & Train/Infer & \xmark & \cmark & \na \\
      & PhysPT~\cite{zhang2024physpt}             & Projection   & Train       & \cmark & \xmark & \na \\
      & Zhang et~al.~\cite{zhang2025plug}         & Projection     & Infer       & \xmark & \cmark & \na \\
      & PhysDiff~\cite{yuan2023physdiff}          & Projection     & Sampling    & \cmark & \cmark & \na \\
    \midrule
    \multirow{4}{*}{\begin{tabular}[c]{@{}l@{}}\emph{Rewards}\\[-1pt]\emph{/ Finetune}\end{tabular}}
      & ReinDiffuse~\cite{han2025reindiffuse}     & Finetuning      & Finetune & \cmark & \cmark & \xmark \\
      & HY-Motion 1.0~\cite{wen2025hy}            & Finetuning      & Finetune & \cmark & \cmark & \xmark \\
      & Morph~\cite{li2025morph}                  & Data refinement & Data     & \cmark & \xmark & \na    \\
      & \ours                                     & Finetuning      & Finetune & \cmark & \cmark & \cmark \\
    \bottomrule
  \end{tabular}
  }
\end{table}

\noindent\textbf{Physics-based character and robot control.}
Physics-based policies have been widely explored in character animation and humanoid robot control. Most existing work~\cite{peng2021amp,peng2018deepmimic,Luo2023PerpetualHC,he2024learning,he2025asap,allshire2025visual} focuses on whole body control (WBC) which is conditioned on the full body target pose and predicts the corresponding plausible action for SMPL~\cite{loper2015smpl} character or humanoid robots through imitation and reinforcement learning. Some works~\cite{tessler2024maskedmimic,luo2024universal,he2024omnih2o,he2025hover,liao2025beyondmimic} further explore versatile humanoid control via policy distillation such as DAgger~\cite{ross2011reduction} to distill a WBC teacher policy to enable partial observation input control. Recent works further extend it to high-level inputs such as text~\cite{tessler2024maskedmimic,wu2025uniphys,he2024omnih2o} and vision~\cite{yin2025visualmimic}. 
One idea is also to perform DAgger~\cite{ross2011reduction} to learn an end-to-end text control policy~\cite{tessler2024maskedmimic,wu2025uniphys}, the other is to leverage a text-driven motion generator to provide target poses~\cite{he2024omnih2o,tevet2024closd,xie2026textop,li2026w1} or latent action~\cite{luo2025sonic,li2025language}, and then perform motion tracking. However, current end-to-end text control policies such as MaskedMimic~\cite{tessler2024maskedmimic} can hardly achieve satisfying text-motion consistency. 
Unlike existing work~\cite{wu2025uniphys,xie2026textop} which leverages short action phrases from BABEL~\cite{punnakkal2021babel} dataset, we focus on natural language descriptions from HumanML3D~\cite{guo2022generating}. Additionally, simply applying motion tracking on noisy motion from the motion generators could fail, existing work~\cite{liao2025beyondmimic,he2024omnih2o} mainly focuses on improving the robustness of whole body controller, while we focus on finetuning generators.

\section{Method}
\label{sec:method}

\begin{figure}[tb]
  \centering
  \includegraphics[width=\textwidth]{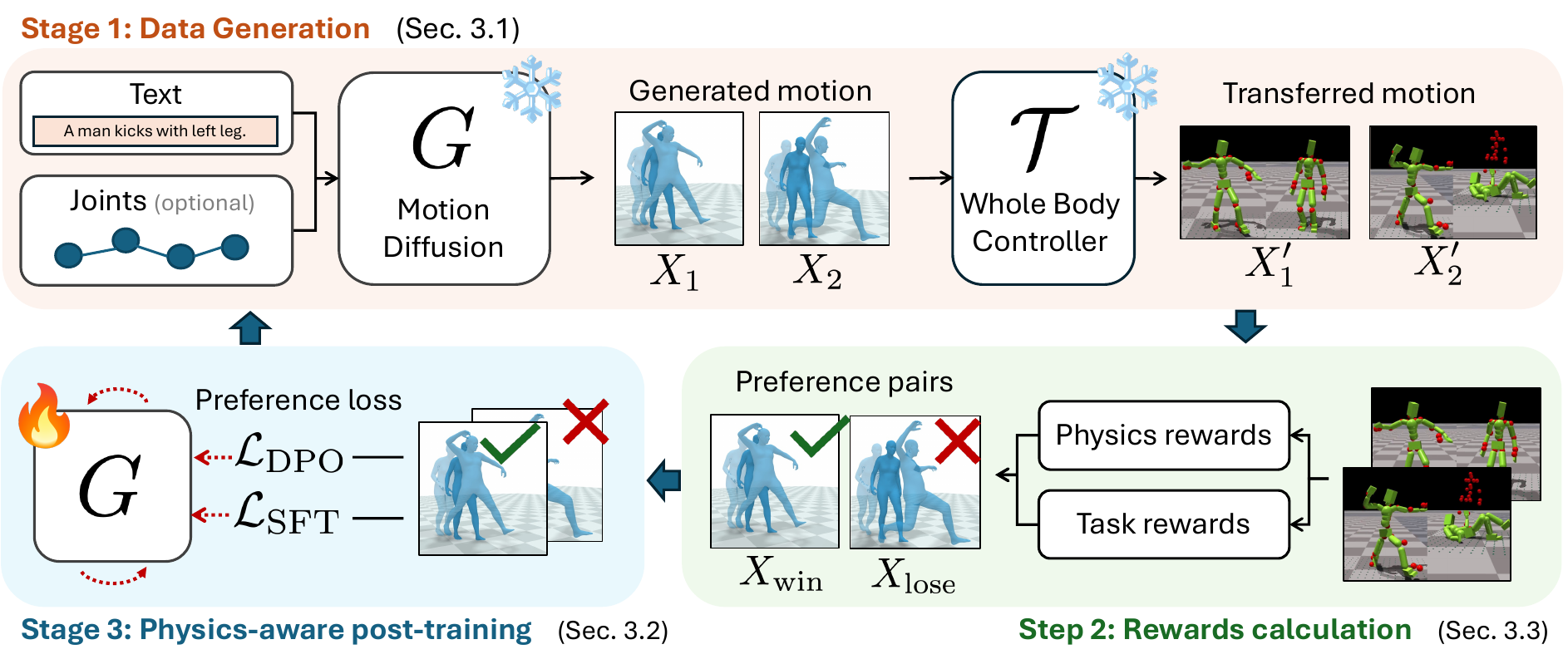}
    \caption{\textbf{Overview of \ours.} Given a conditioning signal (text and optional joint controls), we sample multiple motions $X$ from a pretrained generator.
    A fixed physics-based tracking policy then projects each sample into a simulated trajectory $X'$.
    We compute physics rewards and task rewards on $X'$,
    construct preference pairs, and finetune the generator with DPO. This generation--finetuning procedure can be iterated.}
  \label{fig:method}
\end{figure}

Here, we present \ours. Our core idea is to post-train a motion diffusion model with a physics-grounded reward signal. As shown in Figure~\ref{fig:method}, for each motion generated with a diffusion model, a fixed tracking policy (whole body controller) produces a simulated trajectory.
For each trajectory we then compute physics and task-specific rewards and use them to construct preference pairs for DPO~\cite{rafailov2023direct} finetuning. 
Below we provide details of our method and first highlight the need of custom strategies for deployment-oriented motion generation in Section~\ref{subsec:setup}. In Section~\ref{subsec:dpo}, we formulate the training objectives and describe our post-training procedure based on reinforcement learning. We describe our rewards strategy in Section~\ref{subsec:reward}.

\subsection{Physically-plausible transfer of generated motion}
\label{subsec:setup}

We denote a conditional motion diffusion generator as $G_\theta$. Let $\mathcal{X}_{\text{kin}}\subseteq\mathbb{R}^{N\times D}$ denote the space of kinematic motion sequences in our pose representation, and let $\mathcal{C}$ denote the space of conditioning signals. The generator maps Gaussian noise and a condition to a kinematic motion sample:
\begin{equation}
G_\theta:\ \mathcal{E}\times \mathcal{C}\ \rightarrow\ \mathcal{X}_{\text{kin}}, 
\qquad X = G_\theta(\epsilon, C)\in\mathcal{X}_\text{kin},\ \epsilon\sim\mathcal{N}(0,I),
\label{eq:setup_sampling_typed}
\end{equation}
where $\mathcal{E}$ is the noise space and $\mathcal{C}$ the condition space. We consider two conditioning regimes: (i) text-to-motion, where $C=C_t$ is a text prompt, and (ii) spatially-controlled text-to-motion, where $C=(C_t,C_s)$ and $C_s$ specifies sparse target trajectories for selected joints. Diffusion training makes $G_\theta$ an approximation of the data-generating process for kinematic motions, i.e., samples $X\in\mathcal{X}_{\text{kin}}$ produced by $G_\theta(\epsilon,C)$ match the kinematic motion statistics observed in the dataset. However, this kinematic space does not explicitly enforce physical constraints such as contact consistency, balance, frictional interactions, or actuation limits. We therefore distinguish physically realizable motions under a given embodiment and physics simulator as a subset $\mathcal{X}_{\text{phys}}\subset \mathcal{X}_{\text{kin}}$. Importantly, generating a plausible kinematic motion $X\in \mathcal{X}_{\text{kin}}$ does not imply $X\in \mathcal{X}_{\text{phys}}$: a motion can look realistic and satisfy $C$ in pose space, yet become non-executable once contacts and dynamics are enforced. 
To make the transfer explicit, we introduce a physics operator $\mathcal{T}$ implemented by a fixed tracking controller in a physics simulator, formalized as a mapping between the two spaces:
\begin{equation}
\mathcal{T}:\mathcal{X}_{\text{kin}}\rightarrow \mathcal{X}_{\text{phys}}, 
\qquad X'=\mathcal{T}(X).
\label{eq:setup_tracker}
\end{equation}
We interpret $\mathcal{T}$ as the deployment step: it produces an executable trajectory $X'$ by enforcing dynamics and contact feasibility, but it may modify the reference motion $X$ to satisfy these constraints. See  The magnitude of the required physics-induced correction is measured by the tracking distortion
\begin{equation}
\Delta(X)\ \triangleq\ \|X'-X\|_2^2\ =\ \|\mathcal{T}(X)-X\|_2^2.
\label{eq:setup_distortion}
\end{equation}
Intuitively, if $X$ is already close to $\mathcal{X}_{\text{phys}}$, then $\mathcal{T}$ makes only minor adjustments and $\Delta(X)$ is small; conversely, kinematically plausible but physically inconsistent samples require larger corrections. As a consequence, we argue that for diffusion-based motion generation targeting deployment \textit{it is crucial to move the evaluation target from the kinematic space to deployed space}. In other words, instead of judging a generator by the immediate sample $X$, we judge it by the realized trajectory $X'=\mathcal{T}(X)$ produced by a fixed physics tracker, and we report all quality and adherence metrics on $X'$. This directly captures the transfer failure mode we identify: high-quality kinematic samples can become poor deployed motions when realization induces large corrections, whereas physically compatible samples remain faithful after tracking.
\medskip

\noindent\textbf{Motion setup.}
We represent motions in an SMPL-based format to avoid expensive test-time inverse kinematics required by the commonly used HumanML3D representation. Concretely, we adopt SMPL joint rotations and corresponding kinematic features following recent practice. For models originally trained in HumanML3D format, we retrain them using the SMPL-based representation for a consistent comparison. Unless otherwise specified, $\mathcal{T}$ is instantiated using a fixed pretrained DeepMimic~\cite{peng2018deepmimic} tracker.

\subsection{Physics-aware post-training}\label{subsec:dpo}
Now, our objective is to finetune $G$ to generate kinematic motions that transfer well to simulation after applying $\mathcal{T}$. Since differentiating through $\mathcal{T}$ is challenging, we treat it as a black box and use preference-based post-training. We adopt Direct Preference Optimization (DPO) as our post-training framework. For each condition $C$, we exploit the stochasticity of the diffusion generator to sample $K$ candidate kinematic motions, realize each candidate in simulation, and score the realized motion with a physics-aware reward $\mathcal{R}$, described in the next section:
\begin{equation}
X_k = G_\theta(\epsilon_k, C),\quad \epsilon_k\sim\mathcal{N}(0,I),\qquad
X'_k=\mathcal{T}(X_k),\qquad r_k=\mathcal{R}(X'_k, C).
\label{eq:pref_sampling}
\end{equation}
We then select a preference pair $(X_\text{win}, X_\text{lose})$ by selecting samples from the $K$ candidates and imposing that $X_\text{win}$ achieves a better $\mathcal{R}$ than $X_\text{lose}$. This writes:
\begin{equation}
\begin{aligned}
X_{\text{win}} = X_{k^+},&\qquad X_{\text{lose}} = X_{k^-},
\qquad\mathcal{R}(X'_{k^+},C) \succ \mathcal{R}(X'_{k^-},C),\\\
&k^+,k^- \in \{1,\ldots,K\},\ \ k^+\neq k^-.
\end{aligned}
\end{equation}
Importantly, rewards are computed on the realized motions $X'_k=\mathcal{T}(X_k)$, but the preference pair is defined over the corresponding kinematic samples $(X_{\text{win}},X_{\text{lose}})$, which are the objects used for training. This respects the original DPO formulation, that exploits data directly sampled from the model~\cite{rafailov2023direct}. We optimize the generator using the standard DPO objective from the original paper~\cite{wallace2024diffusion}, which we denote by $\mathcal{L}_{\text{DPO}}(X_{\text{win}},X_{\text{lose}})$. Intuitively, this loss teaches the model to sample motions like $X_{\text{win}}$ more often and motions like $X_{\text{lose}}$ less often under the same condition $C$. In practice, preference-only optimization may drift and harm sample quality, so we add a supervised fine-tuning term on the winning samples only, denoted $\mathcal{L}_{\text{SFT}}(X_{\text{win}})$, to preserve generative capabilities. Overall, our post-training objective is
\begin{equation}
\mathcal{L}=\mathcal{L}_{\text{DPO}}(X_{\text{win}},X_{\text{lose}})+\lambda_{\text{SFT}}\mathcal{L}_{\text{SFT}}(X_{\text{win}}),
\label{eq:total_loss}
\end{equation}
where $\lambda_{\text{SFT}}$ balances preference learning and stabilization. Finally, we apply DPO iteratively: after updating the generator, we resample new candidates and reconstruct preference pairs using the improved model, progressively refreshing the preference data toward the model’s current transfer failure modes under $\mathcal{T}$.

\subsection{Rewards and pair construction}\label{subsec:reward}

We now describe the reward used in Eq.\eqref{eq:pref_sampling}. Specifically, we combine different rewards to achieve best performance. First, we use two \textit{physics rewards} to encourage motions that transfer with minimal correction and stable contacts. Since strict physics following may hurt realism and smoothness of generated motion, we define additional \textit{task rewards} to preserve conditioning and prevent degenerate solutions (e.g., static motion). We first define a tracking reward $\mathcal{R}_\text{track}$, explicitly minimizing the difference between $X$ and $X'$:
\begin{equation}
\mathcal{R}_{\text{track}}(X',X)\triangleq -\Delta(X)= -\|X'-X\|_2^2 .
\label{eq:r_track}
\end{equation}
The second physics reward $\mathcal{R}_\text{slide}$ penalizes foot micro-sliding, which is typically introduced by transfer to $\mathcal{X}_\text{phys}$. Let $h_{\text{feet}}(X',i)$ and $v^{xy}_{\text{feet}}(X',i)$ denote the foot height and horizontal foot speed at frame $i$ computed from the realized motion $X'$. We define
\begin{equation}
\mathcal{R}_{\text{slide}}(X')\triangleq
-\frac{1}{N}\sum_{i=1}^{N}\mathbf{1}\!\left[h_{\text{feet}}(X',i)<h_0\right]\mathbf{1}\!\left[v^{xy}_{\text{feet}}(X',i)>v_0\right],
\label{eq:r_slide}
\end{equation}
which discourages unrealistic contact patterns where feet are near the ground but drift horizontally. For task rewards, we first identify the need to preserve text alignment. We use TMR~\cite{petrovich2023tmr} and define explicitly a metric on text adherence, minimizing the cosine difference of the encoded motion and textual description in a shared latent space:
\begin{equation}
\mathcal{R}_{\text{M2T}}(X',C_t)\triangleq
\cos\!\Big(\mathrm{TMR}_{\text{text}}(C_t),\ \mathrm{TMR}_{\text{mot}}(X')\Big),
\quad
\cos(a,b)\triangleq \frac{\langle a,b\rangle}{\|a\|\,\|b\|}.
\label{eq:r_m2t}
\end{equation}
$\text{TMR}_\text{text}$ and $\text{TMR}_\text{mot}$ refer to the textual and motion encoder of TMR, respectively. When $C$ includes sparse spatial targets $C_s$, we additionally reward matching those targets within $X'$. We define this reward $\mathcal{R}_\text{control} as$:
\begin{equation}
\mathcal{R}_{\text{control}}(X',C_s)\triangleq
-\frac{\|W\odot(X'-C_s)\|_2^2}{\|W\|_1}.
\label{eq:r_control}
\end{equation}
Here, $W$ refers to a joint mask indicating only the control joints available. We combine rewards through a dominance rule when constructing preferences. We define $\mathcal{S}(C)=\{\mathcal{R}_\text{track},\mathcal{R}_\text{slide},\mathcal{R}_\text{M2T}\}$ as the set including rewards conditions for text-to-motion, and $\mathcal{S}(C)=\{\mathcal{R}_\text{track},\mathcal{R}_\text{slide},\mathcal{R}_\text{M2T},\mathcal{R}_\text{control}\}$ when spatial control is provided. We define the composite preference $\mathcal{R}(X',C)$ implicitly by declaring that a realized motion $X'_{k}$ is preferred to $X'_{l}$ under the same condition $C$ if it improves \emph{every} reward term:
\begin{equation}
\mathcal{R}(X'_k,C)\ \succ\ \mathcal{R}(X'_l,C)
\quad \Longleftrightarrow \quad
\mathcal{R}_s(X'_k,C)>\mathcal{R}_s(X'_l,C)\ \ \forall s\in\mathcal{S}(C).
\label{eq:dominance}
\end{equation}
This keeps the preference signal consistent across physics and task objectives without introducing sensitive reward weights.

\section{Experiments}
\label{sec:experiments}

\subsection{Experimental setup}

\noindent\textbf{Datasets.} We generate DPO preference pairs based on text from HumanML3D~\cite{guo2022generating} dataset and then evaluate on the corresponding test set. For spatial control task, we further extract spatial control signals to generate samples and calculate $\mathcal{R}_\text{control}$. To further evaluate the generalization ability of our proposed \ours strategy, we evaluate on the OMOMO~\cite{li2023object} dataset. OMOMO mainly focuses on human-object interaction, hence the text distribution is different. We use it as out-of-distribution test. As mentioned in Section~\ref{subsec:setup}, all the motion data are trained in SMPL-based representation to enable easier adaptation for downstream tasks. Moreover, we filter out motions which require object supports (such as "go upstairs"), as we add physics property in the simulator, and the robot cannot track those motions without the corresponding object.

\begin{table}[tb]
  \caption{\textbf{Evaluation of text-driven human motion generation with SMPL robot simulation on HumanML3D~\cite{guo2022generating} dataset.} We evaluate MaskedMimic~\cite{tessler2024maskedmimic}, MotionStreamer~\cite{xiao2025motionstreamer} and \ours with text-conditioned generation setting as in MotionStreamer~\cite{xiao2025motionstreamer}. The best results are in bold, and the second best results are underlined.
  }
  \label{tab:text_hml3d}
  \centering
  \begin{tabular}{@{}l|cccc|cc@{}}
    \toprule

    \textbf{Method} & \textbf{MM-Dist} $\downarrow$ & \textbf{R@1} $\uparrow$ & \textbf{R@2} $\uparrow$ & \textbf{R@3} $\uparrow$ & \textbf{FID} $\downarrow$ & \textbf{Jerk} $\downarrow$ \\
    \midrule
    Real after simulation & 16.02 & 0.6682 & 0.8326 & 0.8952 & 34.07 & 35.87 \\
    \midrule
    MaskedMimic~\cite{tessler2024maskedmimic} & 19.73 & 0.4134 & 0.5568 & 0.6305 & 73.79 & 66.08 \\
    \midrule
    MotionStreamer~\cite{xiao2025motionstreamer} & \eu{17.17} & \eu{0.5829} & 0.7510 & 0.8310 & \eu{49.14} & \eu{46.75} \\
    SFT & 17.23 & 0.5779 & \eu{0.7651} & \eu{0.8355} & 49.22 & 48.30 \\

    \ours & \eb{16.95} & \eb{0.5853} & \eb{0.7726} & \eb{0.8517} & \eb{48.29} & \eb{43.60} \\
  \bottomrule
  \end{tabular}
\end{table}

\begin{table}[tb]
  \caption{\textbf{Evaluation of spatial-text human motion controllability with SMPL character control.}
  Left: HumanML3D~\cite{guo2022generating}. Right: OMOMO~\cite{li2023object}.
  We evaluate MaskedMimic~\cite{tessler2024maskedmimic}, OmniControl~\cite{xie2023omnicontrol} under two training settings and \ours with cross-control setting as in OmniControl~\cite{xie2023omnicontrol}.
  The best results are in bold, and the second best results are underlined.}
  \label{tab:spatial_smpl_side_by_side}
  \centering
  \small
  \setlength{\tabcolsep}{1.9pt}
  \renewcommand{\arraystretch}{1.05}
  \resizebox{\columnwidth}{!}{
  \begin{tabular}{@{}cl|ccccc|ccccc@{}}
    \multirow{4}{*}{}
      &\multicolumn{1}{c}{}& \multicolumn{5}{c}{\textbf{\textit{HumanML3D}}} & \multicolumn{5}{c}{\textbf{\textit{OMOMO}}}\\ 
      \toprule

      & \textbf{Method}  & \textbf{Err.}$\downarrow$
         & \textbf{MM-Dist}$\downarrow$ & \textbf{R@3}$\uparrow$ & \textbf{FID}$\downarrow$ & \textbf{Jerk}$\downarrow$
         & \textbf{Err.}$\downarrow$
         & \textbf{MM-Dist}$\downarrow$ & \textbf{R@3}$\uparrow$ & \textbf{FID}$\downarrow$ & \textbf{Jerk}$\downarrow$ \\
    \midrule
      & Real after simulation
      & 0.0536 & 3.150 & 0.7607 & 0.98   & 62.96
      & 0.0485 & 6.679 & 0.1576 & 0.59   & 65.49 \\
    \midrule
      & MaskedMimic~\cite{tessler2024maskedmimic}
      & 0.2493 & 5.149 & 0.4932 & 3.99   & 106.49
      & 0.2140 & \eb{6.639} & 0.1081 & 5.99   & 119.52 \\
    \midrule
    \multirow{3}{*}{\rotatebox{90}{\textit{Original}}}
      & OmniControl~\cite{xie2023omnicontrol}
      & 0.1998 & 4.238 & 0.6123 & 5.82   & 115.12
      & 0.3989 & 7.267 & 0.1315 & 20.59  & 161.95 \\
      & SFT
      & 0.1536 & 3.580 & 0.6875 & 2.63   & 92.56
      & 0.2866 & 7.218 & 0.1419 & 13.28  & 130.69 \\
      & \ours
      & 0.1298 & 3.333 & 0.7168 & 0.93   & 62.31
      & 0.1900 & 6.947 & \eb{0.1536} & 3.86   & 101.88 \\
    \midrule
    \multirow{3}{*}{\rotatebox{90}{\textit{Cross}}}
      & OmniControl~\cite{xie2023omnicontrol}
      & \eu{0.0938} & \eu{3.086} & \eb{0.7584} & 0.75   & 64.07
      & 0.1389 & 6.840 & 0.1419 & 3.89   & 88.57 \\
      & SFT
      & 0.0972 & \eb{3.075} & 0.7539 & \eu{0.68}   & \eu{61.22}
      & \eb{0.1319} & 6.875 & 0.1484 & \eu{2.69}   & \eu{84.55} \\
      & \ours
      & \eb{0.0923} & 3.099 & \eu{0.7549} & \eb{0.66} & \eb{58.02}
      & \eu{0.1339} & \eu{6.8245} & \eu{0.1497} & \eb{1.50} & \eb{76.49} \\
    \bottomrule
  \end{tabular}}
\end{table}

\noindent\textbf{Evaluation metrics.} Different from existing work, as we explain in Sec.~\ref{subsec:setup} we evaluate all the model after the whole-body tracking model~\cite{peng2018deepmimic} within simulation to better measure whether the output motion of generators is plausible and whether the character in simulation follows the input condition $C$. The evaluation protocols are mainly adopted from MotionStreamer~\cite{xiao2025motionstreamer} and OmniControl~\cite{xie2023omnicontrol}. The text-motion consistency is evaluated by the Multi-Modal Distance (MM-Dist) and the top-k retrieved accuracy (R@1, R@2, R@3). Frechet Inception Distance (FID) is calculated to quantify the distribution distance between ground truth and the generated motion inside the simulation. For spatial control task, we evaluate the controllability through masked MSE between $C_s$ and $X'$ (Err.). 
The derivative of acceleration (Jerk) is computed to check the smoothness of $X'$.

\noindent\textbf{Implementation details.}
We initialize our pipeline with pretrained MotionStreamer \cite{xiao2025motionstreamer} (for text-to-motion) and OmniControl~\cite{xie2023omnicontrol} (for spatial-text control).
For MotionStreamer~\cite{xiao2025motionstreamer}, we generate 12 samples per training prompt on HumanML3D~\cite{guo2022generating} and perform post-training by updating only the diffusion head, optimized with AdamW~\cite{loshchilov2017decoupled}.
Following MotionStreamer~\cite{xiao2025motionstreamer}, we use the Two-Forward objective as $L_{\mathrm{SFT}}$ on selected win samples. The SFT baseline sets the DPO weight to zero while keeping other settings unchanged.
For spatial-text control, we follow OmniControl~\cite{xie2023omnicontrol}’s cross-control evaluation and sample a random set of control joints during testing. As the original OmniControl~\cite{xie2023omnicontrol} is trained with random one joint per sample, we further re-train it with random number of control joints (cross-control setting) and compare the performance. More details can be found in the appendix.

\begin{figure}[tb]
  \centering
  \includegraphics[width=0.98\textwidth]{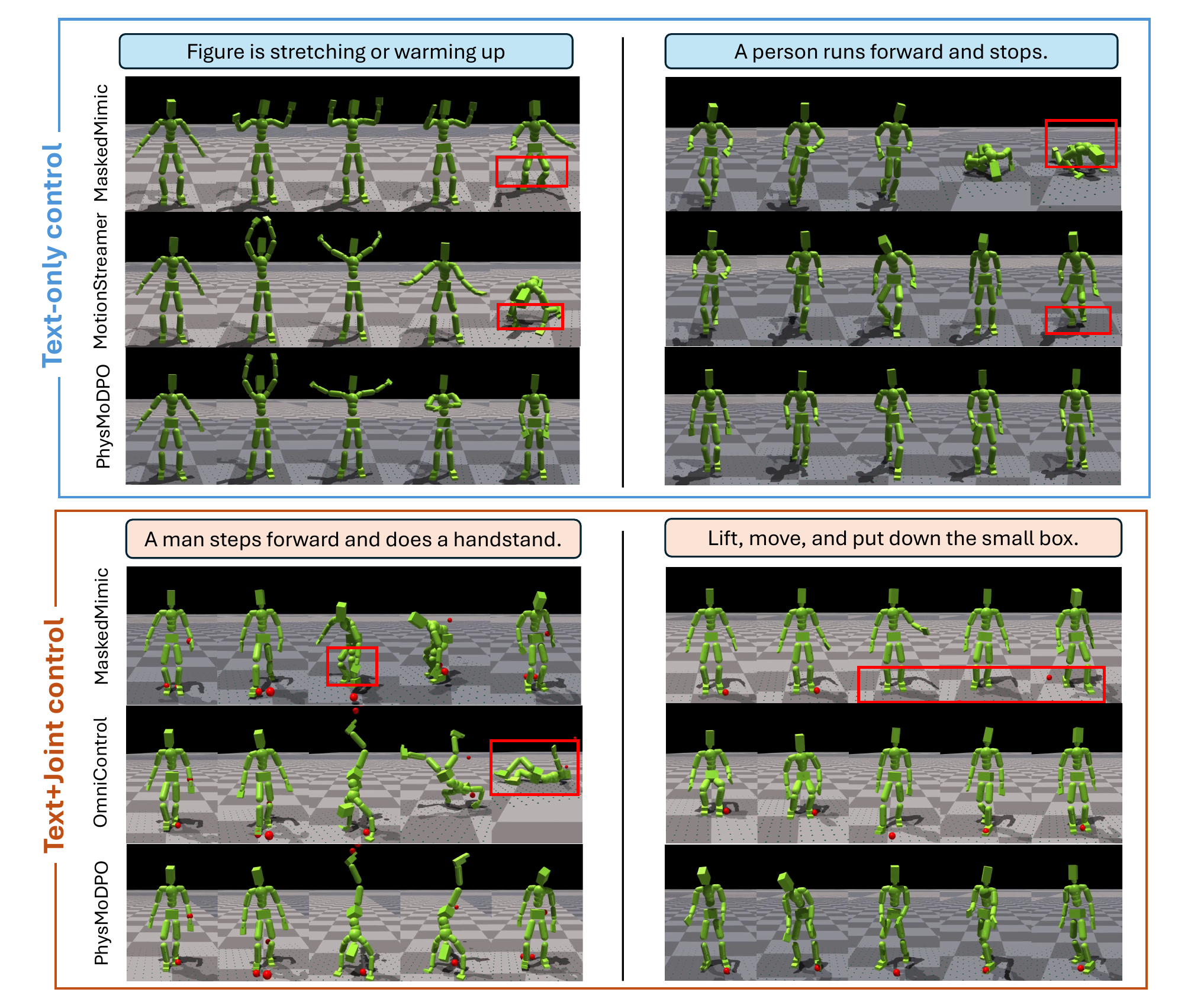}
  \caption{\textbf{Visual comparison with SMPL simulation.} On top, we compare MaskedMimic~\cite{tessler2024maskedmimic}, MotionStreamer~\cite{xiao2025motionstreamer} and \ours on text-to-motion generation task on HumanML3D~\cite{guo2022generating} dataset. At the bottom, we show visual results of spatial-text control task on HumanML3D~\cite{guo2022generating} (left) and OMOMO~\cite{li2023object} (right) dataset. Red balls are the input spatial control signals and red boxes highlight the samples which do not follow the control or lose balance.
  }
  \label{fig:vis_smpl}
\end{figure}

\subsection{Experiments on SMPL simulated character}

\noindent\textbf{Text-to-motion results.}
Table~\ref{tab:text_hml3d} reports text-conditioned motion generation performance on HumanML3D after simulation.
We compare with MaskedMimic~\cite{tessler2024maskedmimic} as physics-based baseline, MotionStreamer~\cite{xiao2025motionstreamer} as our backbone, a naive fine-tuning variant (SFT on win samples), and our proposed method \ours.
Since MaskedMimic directly outputs executable actions in the simulator, hence we omit the projection with tracking policy.
Despite being physically executable, MaskedMimic~\cite{tessler2024maskedmimic} exhibits poor text adherence, \eg, substantially worse retrieval scores (R@1/2/3) and higher FID, indicating limited text-motion consistency.
MotionStreamer~\cite{xiao2025motionstreamer} already achieves strong text-motion alignment after projection, leaving limited room for straightforward supervised refinement.
SFT on win samples yields only marginal gains in text-motion retrieval, while slightly degrading physical fidelity and smoothness.
In contrast, \ours consistently improves text-motion consistency and smoothness.
Concretely, \ours improves text-motion metrics R@3 from 0.8310 to 0.8517 and lowers Jerk from 46.75 to 43.60.
These results demonstrate that our framework effectively reduces the gap between text controllability and physical realism, outperforming both end-to-end physics-based generation and naive fine-tuning.

\noindent\textbf{Spatial-text control results.}
Tables~\ref{tab:spatial_smpl_side_by_side} shows results under the cross-control evaluation protocol.
On HumanML3D~\cite{guo2022generating}, 
\ours consistently improves all aspects over OmniControl~\cite{xie2023omnicontrol} trained on original training setting.
\ours better satisfies multi-joint spatial constraints and also strengthens text-motion consistency while producing higher-quality and smoother motions, as evidenced by a large drop in FID and Jerk. Although OmniControl~\cite{xie2023omnicontrol} trained on cross setting achieves good performance, \ours still improves the spatial controllability and FID.  
On the out-of-distribution OMOMO~\cite{li2023object} dataset, \ours remains robust and delivers significant gains despite not using ground-truth motions as training labels.
Specifically, \ours greatly lowers controllability errors and the generated motions become noticeably more realistic and smoother (lower FID and Jerk).
These results suggest that our strategy generalizes well and can effectively adapt the model under distribution shift.

\noindent\textbf{Qualitative results.}
As presented in Figure~\ref{fig:vis_smpl}, we compare MaskedMimic~\cite{tessler2024maskedmimic}, our finetuning baseline and \ours for text driven motion generation and spatial-text motion generation tasks. From the figure, we can find that MaskedMimic~\cite{tessler2024maskedmimic} can hardly follow the text construction, \eg, in the bottom-left figure, the character does not even try to perform the handstand action. Even though MaskedMimic~\cite{tessler2024maskedmimic} is a physics-based policy trained within a simulation, it may still fall on high-speed motion. As the vanilla diffusion motion generators, \ie MotionStreamer~\cite{xiao2025motionstreamer} and OmniControl~\cite{xie2023omnicontrol}, could generate more implausible motions which cause the SMPL robot losing balance or directly falling down. With our proposed \ours finetuning framework, we have higher probability to generate physical human motions which also follow the input condition. 

\begin{table}[tb]
  \caption{\textbf{Evaluation of text-driven human motion generation with G1 robot on HumanML3D~\cite{guo2022generating} dataset.} We evaluate MaskedMimic~\cite{tessler2024maskedmimic}, MotionStreamer~\cite{xiao2025motionstreamer} and \ours with text-conditioned generation setting as in MotionStreamer~\cite{xiao2025motionstreamer}. The best results are in bold, and the second best results are underlined.
  }
  \label{tab:text_hml3d_g1}
  \centering
  \setlength{\tabcolsep}{4pt}
  \begin{tabular}{@{}l|cccc|cc@{}}
    \toprule
    \textbf{Method} & \textbf{M2T} $\uparrow$ & \textbf{R@1} $\uparrow$ & \textbf{R@2} $\uparrow$ & \textbf{R@3} $\uparrow$ & \textbf{FID} $\downarrow$ & \textbf{Jerk} $\downarrow$ \\
    \midrule
    Real after simulation & 0.8282 & 0.5731 & 0.7583 & 0.8475 & 0.1201 & 87.76 \\
    \midrule
    MaskedMimic~\cite{tessler2024maskedmimic} & 0.7156 & 0.3258 & 0.4762 & 0.5761 & 0.3673 & \eb{83.58} \\
    \midrule
    MotionStreamer~\cite{xiao2025motionstreamer} & \eu{0.7904} & \eu{0.4673} & \eb{0.6620} & \eu{0.7558} & \eu{0.3033} & 95.08 \\
    SFT & 0.7869 & 0.4586 & 0.6444 & 0.7430 & 0.3124 & 97.99 \\
    \ours & \eb{0.7919} & \eb{0.4707} & \eu{0.6596} & \eb{0.7640} & \eb{0.3029} & \eu{90.14} \\
  \bottomrule
  \end{tabular}
\end{table}

\begin{table}[tb]
  \caption{\textbf{Evaluation of human motion controllability with G1 robot.}
  Left: HumanML3D~\cite{guo2022generating}. Right: OMOMO~\cite{li2023object}.
  We evaluate MaskedMimic~\cite{tessler2024maskedmimic}, OmniControl~\cite{xie2023omnicontrol} and \ours with cross-control setting as in OmniControl~\cite{xie2023omnicontrol}.
  The best results are in bold, and the second best results are underlined.}
  \label{tab:spatial_g1_side_by_side}
  \centering
  \small
  \setlength{\tabcolsep}{4pt}
  \renewcommand{\arraystretch}{1.05}
  \resizebox{\columnwidth}{!}{
  \begin{tabular}{cl|ccccc|ccccc}
    \multirow{4}{*}{}
      &\multicolumn{1}{c}{}& \multicolumn{5}{c}{\textbf{\textit{HumanML3D}}} & \multicolumn{5}{c}{\textbf{\textit{OMOMO}}}\\ 
      \toprule

      & \textbf{Method}  & \textbf{Err.}$\downarrow$
         & \textbf{M2T}$\uparrow$ & \textbf{R@3}$\uparrow$ & \textbf{FID}$\downarrow$ & \textbf{Jerk}$\downarrow$
         & \textbf{Err.}$\downarrow$
         & \textbf{M2T}$\uparrow$ & \textbf{R@3}$\uparrow$ & \textbf{FID}$\downarrow$ & \textbf{Jerk}$\downarrow$ \\
    \midrule
      & Real after simulation
      & 0.1894 & 0.7810 & 0.7241 & 0.1287 & 160
      & 0.1730 & 0.4985 & 0.1111 & 0.1099 & 184 \\
    \midrule
      & MaskedMimic~\cite{tessler2024maskedmimic}
      & 0.3388 & 0.6470 & 0.3958 & 0.4626 & 162
      & 0.3157 & 0.4635 & 0.0970 & 0.4324 & 198 \\
    \midrule
    \multirow{3}{*}{\rotatebox{90}{\textit{Original}}}
      & OmniControl~\cite{xie2023omnicontrol}
      & 0.5954 & 0.7025 & 0.5504 & 0.3655 & 315
      & 0.5951 & 0.4773 & 0.1301 & 0.5007 & 459 \\
      & SFT
      & 0.4606 & 0.7335 & 0.6108 & 0.3047 & 250
      & 0.4857 & 0.4836 & 0.1304 & 0.4571 & 369 \\
      & \ours
      & 0.2918 & 0.7414 & 0.6335 & 0.3090 & 142
      & 0.3499 & 0.5011 & 0.1291 & 0.3816 & 239 \\
    \midrule
    \multirow{3}{*}{\rotatebox{90}{\textit{Cross}}}
      & OmniControl~\cite{xie2023omnicontrol}
      & 0.2372 & \eu{0.7830} & 0.7301 & \eb{0.1972} & 156
      & 0.2687 & 0.5164 & 0.1403 & 0.3189 & 192 \\
      & SFT
      & \eu{0.2338} & \eb{0.7834} & \eu{0.7336} & \eu{0.1980} & \eu{155}
      & \eu{0.2545} & \eu{0.5328} & \eu{0.1503} & \eb{0.3055} & \eu{174} \\
      & \ours
      & \eb{0.2240} & 0.7820 & \eb{0.7350} & \eu{0.1980} & \eb{148}
      & \eb{0.2510} & \eb{0.5387} & \eb{0.1571} & \eu{0.3065} & \eb{167} \\
    \bottomrule
  \end{tabular}}
\end{table}

\subsection{Zero-shot transfer to Unitree G1 robot}

\begin{figure}[tb]
  \centering
  \includegraphics[width=0.85\textwidth]{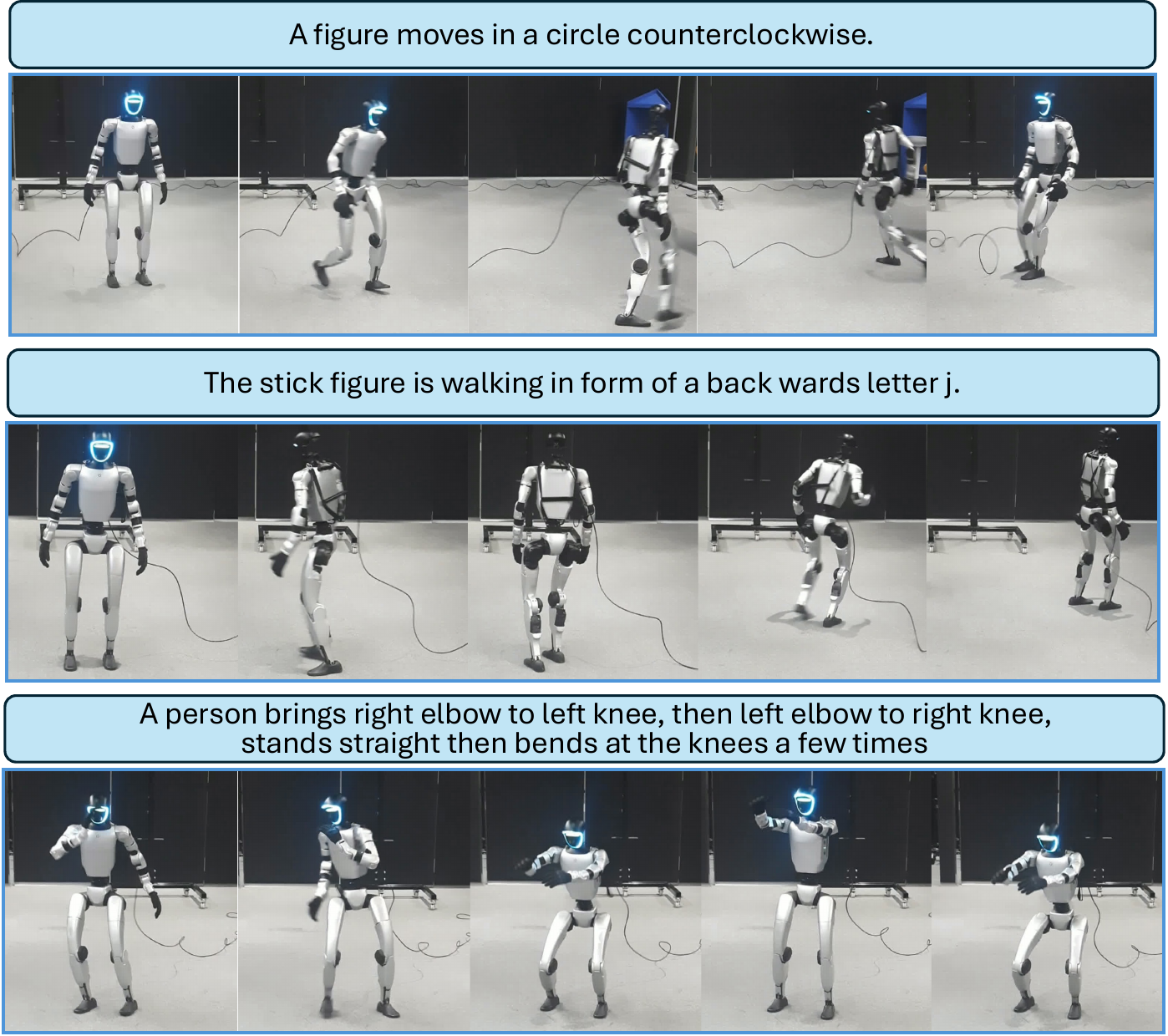}
  \caption{\textbf{Visual results on Unitree G1 robot.} Our deployed motion model enables the robot to move in a physically-realistic way while following input text instructions.
  }
  \label{fig:vis_g1}
\end{figure}

\noindent \textbf{G1 evaluation pipeline.} To show that the motion generated by \ours could better generalize to real robot embodiment, we evaluate our method in a zero-shot manner on Unitree G1 by directly reusing the model trained on SMPL.
Given an input condition, we first run SMPL motion generator to sample motions. Since G1 has different kinematics, we retarget each generated motion to the G1 skeleton before deployment. The retargeted motion is then executed by DeepMimic~\cite{peng2018deepmimic} pretrained on G1.
As for text-motion metrics, 
although the generator is kept unchanged, the standard text-motion metrics depend on the motion representation. Therefore, we retarget AMASS motions from SMPL to G1 and retrain a G1-based TMR~\cite{petrovich2023tmr} evaluator using the same training protocol as the original TMR. This evaluator is only used at inference time for reporting FID and text-motion alignment metrics such as M2T score and top-k retrieval accuracy.
For spatial control, retargeting may rescale the target joint trajectories due to embodiment differences. Thus, when measuring controllability on G1, we compare the simulated G1 motion against the retargeted ground-truth motion on the controlled joints, rather than directly using the original SMPL-space control signal, to avoid mismatched scales while preserving the intended spatial constraints.\smallskip

\noindent\textbf{Text-to-motion results.}
As reported in Table~\ref{tab:text_hml3d_g1}, our SMPL-trained model generalizes to Unitree G1 without any additional training and achieves the strongest overall performance in simulation. The naive SFT trained on SMPL data performs worse than the baseline when zero-shot transferring to the G1 robot. In constrast, our consistent improvement based on MotionStreamer~\cite{xiao2025motionstreamer} proves the effectiveness of the proposed \ours method. Similar to the SMPL character results with simulation, MaskedMimic~\cite{tessler2024maskedmimic} can hardly follow the text input, which results in the lowest text retrieval performance. \smallskip

\noindent\textbf{Spatial-text control results.}
The comparison with MaskedMimic~\cite{tessler2024maskedmimic} and OmniControl~\cite{xie2023omnicontrol} can be found in Table~\ref{tab:spatial_g1_side_by_side}. Compared with the original OmniControl~\cite{xie2023omnicontrol}, \ours produces motions that are noticeably smoother, while maintaining stronger consistency to the spatial control signals. 
Overall, the zero-shot results indicate that our post-training enhances both physical feasibility and control faithfulness, and these gains transfer well across embodiments.\smallskip

\noindent \textbf{Real robot deployment.} As the DeepMimic~\cite{peng2018deepmimic} focuses for on character animation, and can hardly be deployed on real robot, we train a whole body control policy for G1 robot following BeyondMimic~\cite{liao2025beyondmimic}, and then we deploy the policy to execute the retargeted G1 motion from pipeline for real robot experiments.
As shown in Figure~\ref{fig:vis_g1}, we deploy the motions generated by \ours on Unitree G1 robot. The zero-shot examples prove the plausibility of our generated motions.\smallskip

\begin{figure}[t]
  \centering
  \includegraphics[width=\textwidth]{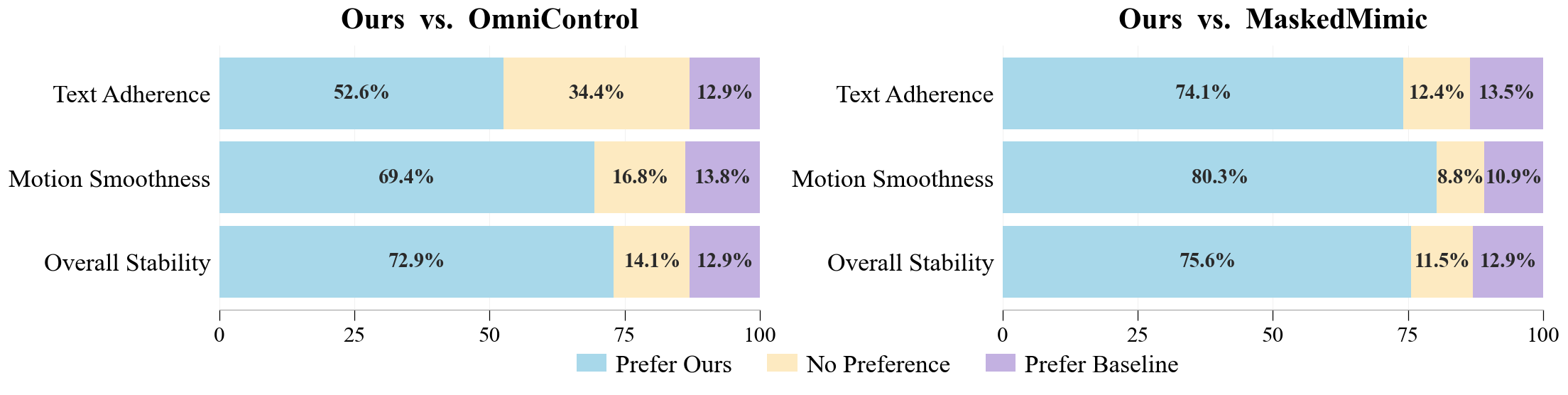}
  \caption{\textbf{User study.} Comparison of  real-robot motion sequences generated by \ours, MaskedMimic~\cite{tessler2024maskedmimic} and OmniControl~\cite{xie2023omnicontrol}.
  \ours outperform both competitors in terms of text adherence, motion smoothness and overall stability.
  }
  \vspace{-0.2cm}
  \label{fig:user_study}
\end{figure}

\noindent \textbf{User study.} 
We conduct a user study to complement metrics.
For each method, we present 40 paired real robot videos to participants together with the input text.
Twenty participants are asked to select videos according to text adherence, motion smoothness and robot stability. As shown in Figure~\ref{fig:user_study}, \ours consistently outperform OmniControl~\cite{xie2023omnicontrol} and MaskedMimic~\cite{tessler2024maskedmimic} across all the criteria. The detailed design of user study can be found in the appendix.

\subsection{Ablation study}
\label{subsec:ablation}

As summarized in Table~\ref{tab:ablation_merge}, we conduct ablations on iterative training round and rewards setting.

\begin{table}[tb]
  \caption{\textbf{Ablation studies.} We train OmniControl~\cite{xie2023omnicontrol} and study the impact of the number of iterative training round and different rewards on HumanML3D~\cite{guo2022generating} dataset. The best results are in bold.}
  \label{tab:ablation_merge}
  \centering

  \begin{subtable}[t]{0.48\textwidth}
    \centering
    \resizebox{0.99\textwidth}{!}{
    \begin{tabular}{@{}l|cccc@{}}
      \toprule
      Number & \textbf{Err.} $\downarrow$ & \textbf{MM-Dist} $\downarrow$ & \textbf{FID} $\downarrow$ & \textbf{Jerk} $\downarrow$ \\
      \midrule
      1 & 0.1421 & 3.36 & 1.17 & 72.13 \\
      2 & 0.1324 & 3.34 & 0.97 & 63.55 \\
      3 & \eb{0.1298} & \eb{3.33} & \eb{0.93} & \eb{62.31} \\
      \bottomrule
    \end{tabular}
    }
    \caption{Number of iterative training round}

  \end{subtable}\hfill
  \begin{subtable}[t]{0.48\textwidth}
    \centering
    \resizebox{0.99\textwidth}{!}{
    \begin{tabular}{@{}l|cccc@{}}
      \toprule
      Rewards & \textbf{Err.} $\downarrow$ & \textbf{MM-Dist} $\downarrow$ & \textbf{FID} $\downarrow$ & \textbf{Jerk} $\downarrow$ \\
      \midrule
      Tracking & 0.1467 & 3.43 & 1.61 & 74.76 \\
      + Control & 0.1447 & 3.41 & 1.45 & 74.47 \\
      + Sliding & 0.1422 & 3.40 & 1.21 & \eb{68.43} \\
      + M2T & \eb{0.1421} & \eb{3.36} & \eb{1.17} & 72.13 \\
      \bottomrule
    \end{tabular}
    }
    \caption{Rewards}
   
  \end{subtable}
  \vspace{-.8cm}

\end{table}

\noindent\textbf{Iterative training round.}
We first study the effectiveness of the proposed multi-round generation. As the number of iterative rounds increases from 1 to 3, we observe consistent improvements on all metrics: MPJPE drops from 0.0456 to 0.0368, controllability error decreases from 0.1421 to 0.1298, MM-Dist improves from 3.36 to 3.33, while FID is significantly reduced from 1.17 to 0.93 and Jerk decreases from 72.13 to 62.31. These results validate that regenerating preference pairs with the newly improved model provides progressively better supervision, leading to better motion realism and stronger physical plausibility.

\noindent\textbf{Rewards.}
We ablate each reward component in the preference construction. Using only the tracking reward is suboptimal, as it tends to favor overly conservative motions, which makes the training diverge faster. Adding the controllability reward yields a modest gain across metrics, confirming that explicitly optimizing task-following helps preserve condition faithfulness. Introducing the sliding reward brings a clear improvement in both FID and Jerk, showing that penalizing foot skating effectively mitigates common contact artifacts. Finally, adding the M2T score further improves text-motion consistency and generation quality. We note that this last addition slightly increases Jerk, because emphasizing semantic correctness encourages more dynamic actions, which may introduce larger accelerations while still improving realism and alignment overall.

\section{Conclusion}

In this paper, we proposed \ours, a physics-guided post-training framework based on DPO to improve the plausibility of diffusion-based human motion generators. 
To bridge the gap between kinematic generation and physics-based execution, \mbox{\ours} automatically constructs preference pairs through a pretrained physics-based tracking policy with physics-oriented and task-specific rewards. 
We optimize the generator with a DPO objective and improve it further within an iterative generate--finetune loop. 
Experiments on text-to-motion and spatial control tasks demonstrate consistent gains of \ours in physical realism and task metrics in simulation. Moreover, zero-shot transfer to Unitree G1 robot indicates the potential of \ours for robotics-oriented motion generation.\medskip

\noindent\textbf{Limitation and future work.}
Despite encouraging results, \ours has several limitations that suggest directions for future research.
(1) Our current setup primarily considers locomotion on flat ground. Extensions of the method to more diverse terrains will further improve transfer to real-world robotics.
(2) Our construction of preference pairs relies on a fixed simulation tracking policy, which can introduce biases. Future work could incorporate human-validated models to reduce evaluator bias.
Overall, we believe our proposed \ours will advance and inspire research in motion generation and embodied AI.

\section*{Acknowledgements}
This work was in part supported by the Institute of Information \& communications Technology Planning \& Evaluation (IITP) grant funded by the Korean government (MSIT) (No. RS-2024-00457882, National AI Research Lab Project), and by the ANR project CorVis (ANR-21-CE230003-01).

%
%
\bibliographystyle{splncs04}
\bibliography{main}

\appendix

\clearpage
\setcounter{page}{1}

\title{Appendix}
\author{}
\institute{}

\titlerunning{PhysMoDPO}
\authorrunning{Y.~Zhang et al.}

\maketitle

In this appendix, we provide:
\begin{itemize}

\item Section A: Experiments on zero-shot transfer with H1 robot.

\item Section B: Further ablation study on data scale, representation, and hyperparameters.

\item Section C: Implementation details

\end{itemize}

\section{Zero-shot transfer on H1 robot}
\label{sec:exp_supp}

\begin{figure}[t]
  \centering
  \includegraphics[width=\textwidth]{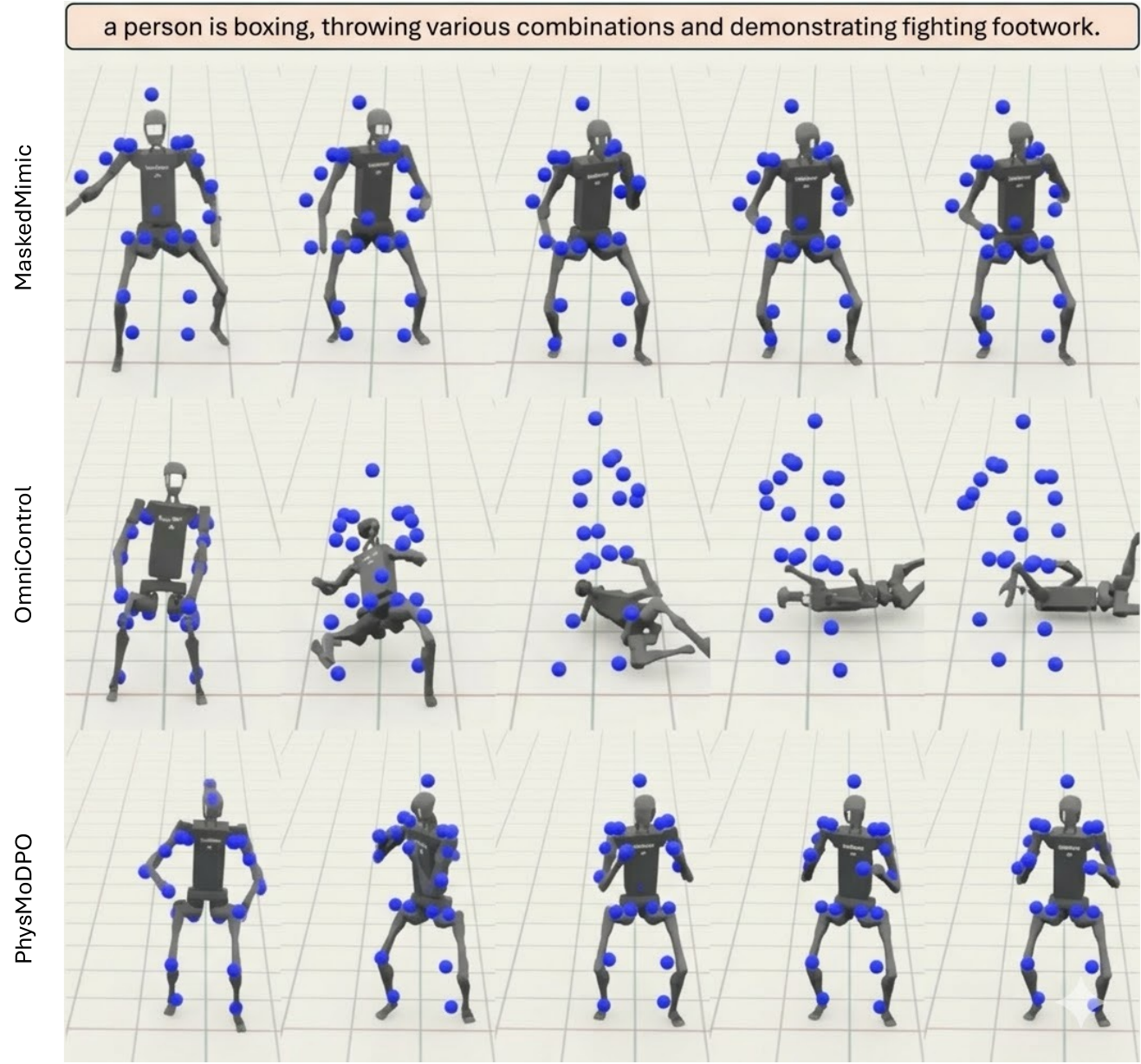}
  \caption{\textbf{Visual comparison with H1 robot.} 
  We transfer the motion generated by  MaskedMimic~\cite{tessler2024maskedmimic}, OmniControl~\cite{xie2023omnicontrol} and \ours to H1 robot and then run Hover~\cite{he2025hover} to track the movement (blue points). See also the website.
  }
  \vspace{-0.2cm}
  \label{fig:vis_h1}
\end{figure}

Similar to the transfer on G1 robot, we pretrain a Whole-Body Controller (WBC) based on HOVER~\cite{he2025hover} on H1 format AMASS~\cite{mahmood2019amass} dataset. Then we directly retarget the output motion from the motion generators and run HOVER~\cite{he2025hover} to obatain the motions under physics constraints. We evaluate and compare the performance between MaskedMimic~\cite{tessler2024maskedmimic}, OmniControl~\cite{xie2023omnicontrol}, SFT baseline and \ours on spatial-text control task in Table~\ref{tab:spatial_hml3d_h1_zero_shot} on HumanML3D~\cite{guo2022generating} test set. Overall, \ours achieves the best performance across spatial controllability metrics, indicating more reliable tracking under the embodiment and controller shift. Meanwhile, \ours maintains competitive text consistency, and also improves motion quality and smoothness. These results demonstrate the robustness of \ours for spatial-text controlled motion generation in zero-shot transfer to H1. 
Additionally, we present visual comparison between MaskedMimic~\cite{tessler2024maskedmimic}, OmniControl~\cite{xie2023omnicontrol} and \ours in Figure~\ref{fig:vis_h1}. We find that instead of boxing, the robot from MaskedMimic~\cite{tessler2024maskedmimic} randomly moves hands, while the motion from OmniControl~\cite{xie2023omnicontrol} gives unstable initial rotation, which causes the robot to fall at the very beginning.

\begin{table}[tb]
  \caption{\textbf{Evaluation of zero-shot human motion controllability with H1 robot on HumanML3D~\cite{guo2022generating} dataset.} We apply the models trained with SMPL simulation and then perform zero-shot evaluation for Unitree H1 robot. The best results are in bold, and the second best results are underlined. 
  }
  \label{tab:spatial_hml3d_h1_zero_shot}
  \centering
  \resizebox{0.99\textwidth}{!}{
  \begin{tabular}{@{}l|ccc|cc|cc@{}}
    \toprule
    \multirow{2}{*}{\textbf{Method}} & \multicolumn{3}{c|}{\textbf{Spatial controllability}} & \multicolumn{2}{c|}{\textbf{Text consistency}} & \multirow{2}{*}{\textbf{FID} $\downarrow$} & \multirow{2}{*}{\textbf{Jerk} $\downarrow$} \\
     & \textbf{Err.} $\downarrow$ & \textbf{Traj err 0.5} $\downarrow$ & \textbf{Traj err 0.2} $\downarrow$ & \textbf{M2T} $\uparrow$ & \textbf{R@3} $\uparrow$ & & \\
    \midrule
    Real after simulation & 0.2406 & 0.2734 & 0.8105 & 0.5915 & 0.3916 & 0.7976 & 203.6 \\
    \midrule
    MaskedMimic~\cite{tessler2024maskedmimic} & 0.5173 & \eu{0.5752} & 0.9629 & 0.5459 & 0.2227 & 1.132 & \eu{191.4} \\
    \midrule
    OmniControl~\cite{xie2023omnicontrol} & 0.5121 & 0.7129 & 0.9639 & 0.5693 & 0.2803 & 1.038 & 279.3 \\
    SFT & \eu{0.3504} & 0.6113 & \eu{0.9443} & \eb{0.5846} & \eu{0.3574} & \eu{0.9523} & 231.4 \\
    \ours & \eb{0.2497} & \eb{0.4951} & \eb{0.9141} & \eu{0.5830} & \eb{0.3643} & \eb{0.8820} & \eb{172.7} \\
  \bottomrule
  \end{tabular}
  }
\end{table}

\section{Ablation study}
\label{sec:ablation_supp}

\noindent\textbf{Data representation.}
As described in Section 3.1 of the main paper, the original HumanML3D data format requires expensive inverse kinematics to convert to SMPL~\cite{loper2015smpl} parameters, which are widely used for further rendering and tracking policy. Therefore, we retrain OmniControl~\cite{xie2023omnicontrol} with SMPL-based representation~\cite{petrovich2024multi}, and compare with the original model in Table~\ref{tab:spatial_hml3d_data_rep}.
We find that switching OmniControl~\cite{xie2023omnicontrol} to the SMPL-based representation consistently improves spatial controllability across all controls and FID.
Although text alignment slightly degrades, given that downstream rendering and tracking operate in SMPL space, we adopt the SMPL representation in the all experiments.

\begin{table}[tb]
  \caption{\textbf{Ablation of data representation.} We compare OmniControl~\cite{xie2023omnicontrol} model trained on different data representations. Numbers are calculated without applying tracking policy. The best results are in bold.}
  \label{tab:spatial_hml3d_data_rep}
  \centering
  \resizebox{0.99\textwidth}{!}{
  \begin{tabular}{@{}l|l|ccc|cc|cc@{}}
    \toprule
    \multirow{2}{*}{\textbf{Control}} & \multirow{2}{*}{\textbf{Data rep}} & \multicolumn{3}{c|}{\textbf{Spatial controllability}} & \multicolumn{2}{c|}{\textbf{Text consistency}} & \multirow{2}{*}{\textbf{FID} $\downarrow$} & \multirow{2}{*}{\textbf{Jerk} $\downarrow$} \\
     &  & \textbf{Err.} $\downarrow$ & \textbf{Traj err 0.5} $\downarrow$ & \textbf{Traj err 0.2} $\downarrow$ & \textbf{MM-Dist} $\downarrow$ & \textbf{R@3} $\uparrow$ & & \\
     \midrule
   \multirow{2}{*}{Pelvis} & HML3D & 0.0724 & 0.1055 & 0.2920 & \eb{3.0137} & 0.7900 & 0.3619 & \eb{33.30} \\
    & SMPL & \eb{0.0352} & \eb{0.0420} & \eb{0.1855} & 3.0255 & \eb{0.7822} & \eb{0.2107} & 34.38 \\
    \midrule
   \multirow{2}{*}{Left hand} & HML3D & 0.1320 & 0.2197 & 0.6504 & \eb{2.9725} & \eb{0.8037} & 0.3053 & \eb{30.92} \\
    & SMPL & \eb{0.0662} & \eb{0.1094} & \eb{0.4756} & 3.0054 & 0.7842 & \eb{0.1881} & 33.37 \\
    \midrule
   \multirow{2}{*}{Right hand} & HML3D & 0.1317 & 0.2188 & 0.6572 & \eb{2.9093} & 0.7939 & 0.2523 & \eb{31.60} \\
    & SMPL & \eb{0.0656} & \eb{0.1006} & \eb{0.4766} & 3.0276 & \eb{0.7656} & \eb{0.1753} & 34.29 \\
  \bottomrule
  \end{tabular}
  }
\end{table}

\begin{table}[tb]
  \caption{\textbf{Ablation studies.} We train OmniControl~\cite{xie2023omnicontrol} and study the impact of the data scale and preference pair selection strategies on HumanML3D~\cite{guo2022generating} dataset. The best results are in bold.}
  \label{tab:ablation_merge_supp}
  \centering
  \begin{subtable}[t]{0.48\textwidth}
    \centering
    
    \resizebox{0.99\textwidth}{!}{
    \begin{tabular}{@{}l|cccc@{}}
      \toprule
      \textbf{Ratio}  & \textbf{Err.} $\downarrow$ & \textbf{MM-Dist} $\downarrow$ & \textbf{FID} $\downarrow$ & \textbf{Jerk} $\downarrow$ \\
      \midrule
      OmniControl~\cite{xie2023omnicontrol} & 0.1998 & 4.23 & 5.82 & 115.12 \\ 
      20 \% & 0.1434 & 3.46 & 1.34 & \eb{67.69} \\
      50 \% & 0.1430 & 3.39 & 1.24 & 70.28 \\
      100 \% & \eb{0.1421} & \eb{3.36} & \eb{1.17} & 72.13 \\
      \bottomrule
    \end{tabular}
    }
    \caption{Different data scales}
  \end{subtable}\hfill
  \begin{subtable}[t]{0.48\textwidth}
    \centering

    \resizebox{0.99\textwidth}{!}{
    \begin{tabular}{@{}l|cccc@{}}
      \toprule
      \textbf{Method}  & \textbf{Err.} $\downarrow$ & \textbf{MM-Dist} $\downarrow$ & \textbf{FID} $\downarrow$ & \textbf{Jerk} $\downarrow$ \\
      \midrule
      OmniControl~\cite{xie2023omnicontrol} & 0.1998 & 4.23 & 5.82 & 115.12 \\ 
      Fuse score & 0.1476 & 3.45 & 1.61 & 78.95 \\
      Dominance & \eb{0.1421} & \eb{3.36} & \eb{1.17} & \eb{72.13} \\
      \bottomrule
    \end{tabular}
    }
    \caption{Pair selection strategies}
  \end{subtable}

\end{table}

\noindent\textbf{Data scale.}
We further evaluate the impact of training data scale by using 20\%, 50\%, and 100\% of the preference pairs. Even with only 20\% data, the model remains reasonably strong for all metrics, indicating that our preference construction is sample-efficient. Increasing the scale consistently improves generation quality (FID) and text-motion alignment (MM-Dist). Interestingly, smaller scales yield slightly lower Jerk, while full data achieves the best overall FID and text-motion consistency with slightly higher dynamics. We attribute this to larger-scale training encouraging more diverse and semantically expressive motions, which may contain stronger accelerations.

\noindent\textbf{Pair selection strategy.}
We compare two strategies for constructing preference pairs from multiple rewards: (i) score fusion by normalizing rewards and using weighted summation (Fuse score), and (ii) our strict dominance-based selection (Dominance), which requires the winning sample to outperform the losing one on all rewards. Weighted fusion is highly sensitive to reward weights and tends to introduce reward engineering and reward hacking. Empirically it performs notably worse. In contrast, the proposed dominance-based selection consistently yields better overall performance and more stable training, demonstrating its effectiveness for multi-objective preference construction without tuning fragile weights.

\noindent\textbf{Ablation on SFT loss ratio $\lambda_{\text{SFT}}$.}
Table~\ref{tab:ablation_parameter} (top) studies the effect of adding an SFT loss on the preferred (win) samples. When $\lambda_{\text{SFT}}=0$, the model shows degraded controllability and generation quality. Increasing $\lambda_{\text{SFT}}$ consistently improves spatial controllability, text consistency, as well as FID and jerk, and reaches the best overall performance at $\lambda_{\text{SFT}}=2$. Further increasing $\lambda_{\text{SFT}}$ (e.g., 5 or 10) yields slight regressions, suggesting overly strong SFT regularization may weaken the benefit of preference optimization. We therefore use $\lambda_{\text{SFT}}=2$ in all experiments.

\noindent\textbf{Ablation on DPO temperature $\beta$.}
Table~\ref{tab:ablation_parameter} (bottom) evaluates the DPO temperature $\beta$ with $\lambda_{\text{SFT}}=2$. A small temperature ($\beta=1$) provides limited improvement, while a moderate value ($\beta=20$) achieves the best overall trade-off across controllability, text alignment, and generation quality. When $\beta$ becomes too large ($\beta=50$), the performance drops significantly (e.g., worse FID and higher jerk), indicating overly aggressive preference updates can harm distributional quality and motion smoothness. Hence, we set $\beta=20$ by default.

\begin{table}[tb]
  \caption{\textbf{Ablation study on hyperparameters.} Here $\lambda_{\text{SFT}}$ is the weight of SFT loss on win sample, $\beta$ is the temperature parameter from Diffusion-DPO~\cite{wallace2024diffusion}. The best results are in bold, and the second best results are underlined.}
  \label{tab:ablation_parameter}
  \centering
  \begin{tabular}{@{}c|c|ccc|cc|cc@{}}
    \toprule
    \multirow{2}{*}{\textbf{$\lambda_{\text{SFT}}$}} & \multirow{2}{*}{\textbf{$\beta$}} & \multicolumn{3}{c|}{\textbf{Spatial controllability}} & \multicolumn{2}{c|}{\textbf{Text consistency}} & \multirow{2}{*}{\textbf{FID} $\downarrow$} & \multirow{2}{*}{\textbf{Jerk} $\downarrow$} \\
     &  & \textbf{Err.} $\downarrow$ & \textbf{Traj err 0.5} $\downarrow$ & \textbf{Traj err 0.2} $\downarrow$ & \textbf{MM-Dist} $\downarrow$ & \textbf{R@3} $\uparrow$ & & \\
    \midrule
   0 & 20 & 0.2348 & 0.5918 & 0.9023 & 4.21 & 0.6133 & 5.28 & 116.77 \\
   1 & 20 & 0.1771 & 0.5469 & 0.8887 & 4.00 & 0.6387 & 4.31 & 103.64 \\
   2 & 20 & \eb{0.1421} & \eb{0.4238} & \eb{0.8418} & \eb{3.36} & \eb{0.7246} & \eb{1.17} & \eb{72.13} \\
   5 & 20 & \eu{0.1440} & \eu{0.4463} & \eu{0.8496} & \eu{3.40} & \eu{0.7139} & \eu{1.82} & \eu{81.11} \\
   10 & 20 & 0.1476 & 0.4629 & 0.8662 & 3.49 & 0.7041 & 2.29 & 88.07 \\
   \midrule
   
    2 & 1 & 0.1513 & 0.4775 & 0.8701 & 3.53 & 0.6973 & 2.42 & 91.31 \\
    2 & 5 & \eu{0.1458} & \eu{0.4492} & \eu{0.8604} & \eu{3.44} & \eu{0.7002} & \eu{1.95} & \eu{85.83} \\
    2 & 20 & \eb{0.1421} & \eb{0.4238} & \eb{0.8418} & \eb{3.36} & \eb{0.7246} & \eb{1.17} & \eb{72.13} \\
    2 & 50 & 0.1848 & 0.5566 & 0.8926 & 4.02 & 0.6367 & 4.41 & 106.02 \\

  \bottomrule
  \end{tabular}
\end{table}

\section{Implementation details}
\label{sec:implementation}

In Section 4.1 of the main paper, we provide a brief overview of the implementation details. For completeness and reproducibility, we present the full experimental configuration in this appendix for text-to-motion and spatial-text control tasks. In addition, we complement the metrics and the user study details.

\noindent\textbf{Implementation details for the text-driven task.} 
We apply the pretrained model from MotionStreamer~\cite{xiao2025motionstreamer} to initialize the data generation pipeline and then generate 12 samples for each text prompt from the training set of HumanML3D~\cite{guo2022generating}. As for the post-training stage,  we fix all the parameters except the diffusion head inside the MotionStreamer. Following the Transformer training setup of MotionStreamer~\cite{xiao2025motionstreamer}, we replace the standard diffusion training losses with Two-Forward strategy as our $L_{SFT}$ loss on win samples $X_{win}$. We train the model using a batch size of 32, with AdamW optimizer~\cite{loshchilov2017decoupled} with learning rate 1e-6 for 5,000 iterations, among which the first 500 steps are warm-up stage. 
The hyperparamters $\lambda_{\text{SFT}}$ and Diffusion-DPO~\cite{wallace2024diffusion} temperature $\beta$ in \ours framework are set to 1 and 5, respectively. As for SFT baseline, we set the DPO loss weight to be zero, while the other hyperparameters remain unchanged. Following OmniControl~\cite{xie2023omnicontrol}, the threshold $h_0$ and $v_0$ in $R_{\text{slide}}$ are set to 0.05~m and 0.50~m/s. Note that, as we remove motions which requires object support for both training and inference, we reproduce all the baseline results with the new filtered test set for fair comparison.

\noindent\textbf{Implementation details for spatial-text task.}
To complement to implementation details in Section 4.1 of the main paper, we further describe the training settings, hyperparameters and baselines.
We follow the training protocol of OmniControl~\cite{xie2023omnicontrol} and evaluate cross-control as in OmniControl, where we randomly sample a variable number of control joints for each test sample. We additionally report single-joint control evaluation in Section B.
We convert HumanML3D~\cite{guo2022generating} into a SMPL-based representation and re-train the OmniControl backbone on this format. An ablation on the data representation is provided in Section B.
The original OmniControl training selects a single control joint per sample, which can mismatch the cross-control evaluation.
Therefore, we additionally re-train OmniControl with the cross-control protocol, i.e., sampling a random set of control joints during training (cross).
We then apply \ours to finetune the cross-trained OmniControl for 4,000 steps with batch size 64, using a linear warm-up over the first 200 steps.
We fix $\lambda_{\text{SFT}}{=}2$ and $\beta{=}20$ for all experiments on spatial-text control task and the corresponding hyperparameter ablations can be found in the Section B.
Beyond HumanML3D, we further adapt the model pretrained on HumanML3D~\cite{guo2022generating} data to the out-of-distribution OMOMO~\cite{li2023object} dataset without using its ground-truth motions as supervised training targets, while keeping the same pipeline and hyperparameters. In terms of MaskedMimic, for fair comparison, we initialize the robot with standard pose and then add 1 second  warm up stage to make the robot move to the first frame spatial control signals, and then we remove the first 30 frames to calculate the metrics.

\noindent\textbf{Additional metrics.} 
For spatial control task, in addition to masked MSE between $C_s$ and $X'$ (Err.), we further evaluate the controllability through failure rate (Traj err) under given threshold. Specifically, Traj err 0.5 represents the ratio of failure motion sequences whose maximum joint error across the whoe sequence is larger than 0.5 meters. 

\begin{figure}[t]
  \centering
  \includegraphics[width=\textwidth]{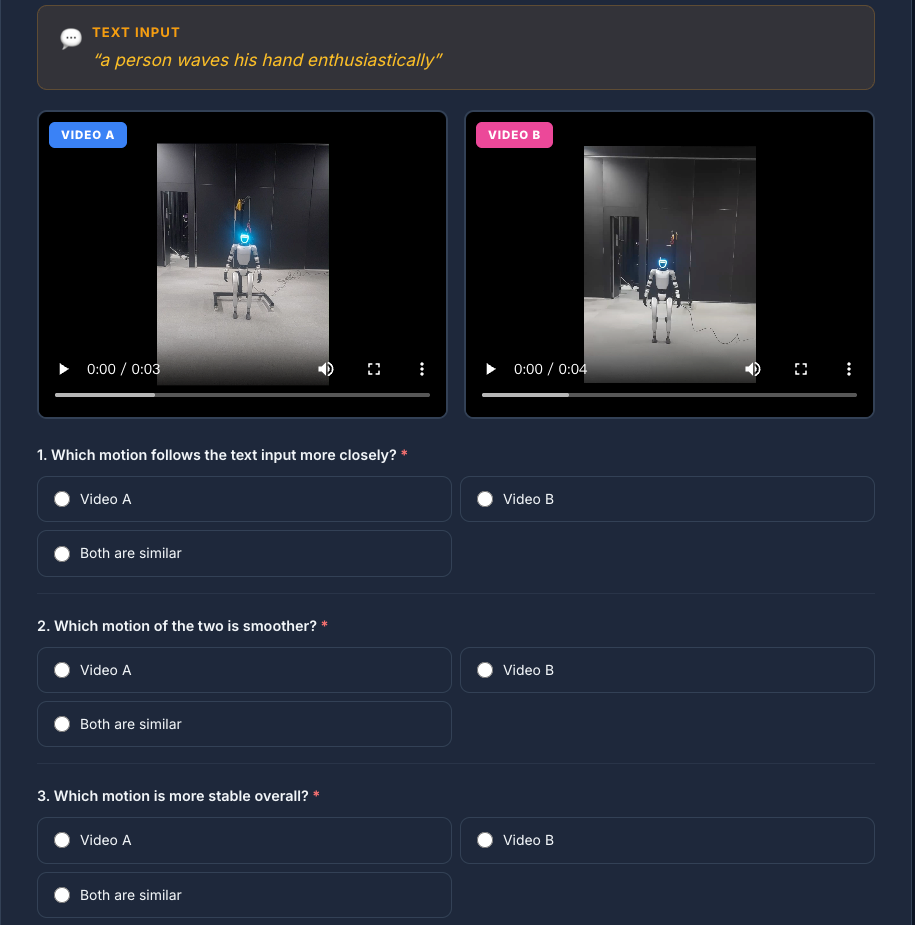}
  \caption{\textbf{User study design.} We show the text input with 2 videos generated by \ours and one baseline method. The participant is asked to evaluate the videos according to text-motion similarity, smoothness and stability.
  }
  \vspace{-0.2cm}
  \label{fig:user_study_design}
\end{figure}

\noindent\textbf{User study details.} We generate 40 videos per method and thus provide 80 questions. As shown in Figure~\ref{fig:user_study_design}, for each question, we show two real robot videos from \ours and one of the baseline methods. Twenty participants evaluated the videos according to 3 aspects: text-motion consistency, smoothness, and stability. For each sub-question, the participant could select one of the videos or "Both are similar".

\end{document}